\pdfoutput=1

\documentclass[11pt]{article}

\usepackage[]{emnlp2021} 

\usepackage{times}
\usepackage{latexsym}

\usepackage[T1]{fontenc}

\usepackage[utf8]{inputenc}

\usepackage{microtype}

\usepackage{mathtools}
\usepackage{amsmath}
\usepackage{multirow}
\usepackage{tabularx}
\usepackage{booktabs}
\usepackage{color}

\usepackage{times}
\usepackage{latexsym}
\usepackage{graphicx}
\usepackage{pgfplots}
\usepackage{amssymb}
\usepackage[ruled]{algorithm2e}
\usepackage{url}
\usepackage{amsfonts}
\usepackage{bm}
\DeclareMathOperator*{\softmax}{softmax}
\DeclareMathOperator*{\pooling}{pooling}
\DeclareMathOperator*{\Conv}{Conv}

\title{Unsupervised Pre-training with Structured Knowledge for\\ Improving Natural Language Inference}

\author{
Xiaoyu Yang~~~~Xiaodan Zhu~~~~Zhan Shi~~~~Tianda Li \\ 
  ECE Department, Queen's University, Canada \\ 
  \texttt{xiaoyu.yang@queensu.ca} \quad 
}

\begin{document}
\maketitle
\begin{abstract}
While recent research on natural language inference has considerably benefited from large annotated datasets~\cite{williams2017broad,snli:emnlp2015}, the amount of inference-related knowledge (including commonsense) provided in the annotated data is still rather limited. There have been two lines of approaches that can be used to further address the limitation: (1) unsupervised pretraining can leverage knowledge in much larger unstructured text data; (2) structured (often human-curated) knowledge has started to be considered in neural-network-based models for NLI. An immediate question is whether these two approaches complement each other, or how to develop models that can bring together their advantages. In this paper, we propose models that leverage structured knowledge in different components of pre-trained models. Our results show that the proposed models perform better than previous BERT-based state-of-the-art models. Although our models are proposed for NLI, they can be easily extended to other sentence or sentence-pair classification problems. 

\end{abstract}

\section{Introduction}
Natural language inference (NLI), also known as recognizing textual entailment (RTE) ~\cite{snli:emnlp2015,dagan2013recognizing, maccartney2008modeling}, is a challenging problem in natural language understanding. It also acts as a test bed for representation learning for natural language~\cite{snli:emnlp2015,williams2017broad, wang2018glue}. Specifically, NLI asks a system to identify the relationship between a premise~\textit{p} and a hypothesis ~\textit{h} such as entailment and contradiction.

Recent advance on NLI has benefited from the availability of large annotated data~\cite{williams2017broad,snli:emnlp2015}. It is, however, unlikely that the annotated data will contain all needed inference knowledge. 

The most recent year has seen two lines of approaches that can be used to help address the limitation. First, the state-of-the-art unsupervised pre-training has shown to be very effective in leveraging knowledge in large unstructured data~\cite{devlin2018bert,radford2018improving, Peters:2018}. In parallel, some research has started to incorporate structured knowledge~\cite{chen2018neural} into neural-network-based NLI models. 
Whether these two approaches complement each other in leveraging knowledge in large unstructured text and structured knowledge. How to develop models  that can bring together the advantages of these two approaches? 

In this paper, we explore a variety of approaches to leverage structured external knowledge in Transformer-based pre-training architectures~\cite{vaswani2017attention}, which are widely used in the state-of-the-art pre-trained models such as generative pre-trained Transformer (GPT) model~\cite{radford2018improving} and BERT~\cite{devlin2018bert}. Specifically we
(1) incorporate structured knowledge in Transformer self-attention,
(2) add a knowledge-specific layer over the existing Transformer block, and (3) incorporate structured knowledge in global inference. We believe that complicated NLP tasks such as NLI can benefit from knowledge available outside the training data from these two typical sources: the knowledge learned implicitly from unstructured text through unsupervised pre-training and that from a structured (often human-created) knowledge base.

We evaluate the proposed models on the widely used NLI benchmarks, including the Stanford Natural Language Inference (SNLI) corpus~\cite{snli:emnlp2015}, Multi-Genre Natural Language Inference (MultiNLI) corpus~\cite{williams2017broad}, and a newly introduced diagnostic Glockner dataset~\cite{glockner2018breaking}. Our results show that the proposed models perform better on these datasets than previous BERT-based state-of-the-art models. While our models are proposed for NLI, they can be easily adapted to other sentence or sentence-pair classification problems. 

\section{Related Work}
\label{sec:relatedwork}
\subsection{Pre-training in NLI and Other Tasks}
Unsupervised pre-trained models~\cite{Peters:2018, radford2018improving, devlin2018bert, radford2019language, song2019mass} have proven to be effective in achieving great improvements in many NLP tasks. Feature-based and finetune-based are two strategies where the pre-trained models apply to downstream tasks. Feature-based models such as ELMo~\cite{Peters:2018} provide pre-trained representations as additional features to original task-specific models. Finetune-based models such as Generative pre-trained Transformer (GPT)~\cite{radford2018improving} and BERT~\cite{devlin2018bert}, however, use pre-trained architectures and then fine-tune the same architectures on downstream tasks. 
In this paper, we explore combining structured knowledge into pre-trained models in the stage of fine-tuning.

\subsection{Human-Authorized External Knowledge Integration}
As discussed above, while neural networks have been shown to be very effective in modeling NLI with large amounts of training data, their end-to-end training is based on a strong assumption that all necessary knowledge for inference is learnable from the provided training data. This kind of assumption, however, can be limited in some circumstances where the knowledge available in a human-authorized external knowledge base is too sparse to learn just from the training data or corpora the model is pre-trained with. 

Existing work has made an effort to incorporate human-authorized knowledge ({\em e.g.}, WordNet, ConceptNet) in neural networks. 
For entity and relation representation of knowledge graph, there are both structure-based and semantically-enriched approaches to get their embeddings. For example, TransE~\cite{bordes2011learning} provides structure-based embedding by modeling relations as translations, while for Neural Tensor Network~\cite{socher2013reasoning}, it represents entities as an average of their word vectors and then it is used for tensor-based transformation. 
After getting vector representation for entities or relations, they can be used to optimize sentence alignment like~\cite{chen2018neural}. Some efforts have also been made to generate knowledge enhanced representations for premise-hypothesis pairs and then used to benefit the final classification like~\cite{wang2018improving}.

\subsection{Evaluation of Models for NLI}
Previous research has made contributions to large annotated datasets for NLI evaluation. For example, SNLI~\cite{snli:emnlp2015} and MultiNLI~\cite{williams2017broad} are two widely used NLI datasets.
Previous research ~\cite{gururangan2018annotation, naik2018stress} has also paid attention to whether existing NLI systems have learned NLI-related semantics or just explored the regularities existing in the data that are not relevant to NLI.

To investigate this, different methods have been proposed.~\cite{wang2018simply} introduces a \textit{swapping} evaluation method, which means changing the distribution of words by swapping a premise with its corresponding hypothesis to test the robustness of models. Also, new test datasets are proposed, {\em e.g.}, Glockner test set~\cite{glockner2018breaking}. In the Glockner test dataset, premises are taken from the SNLI training set, and hypotheses are generated by replacing a single word in its corresponding premise sentence. In addition, swapping Glockner test in \cite{li2019several} shows that external knowledge from unsupervised pre-training and human-authorized external knowledge can benefit models by improving the capacity of capturing NLI semantics. Also, human-authorized external knowledge has the potential to enhance pre-trained models.

\begin{figure*}[!htpb]
\includegraphics[width=16cm]
{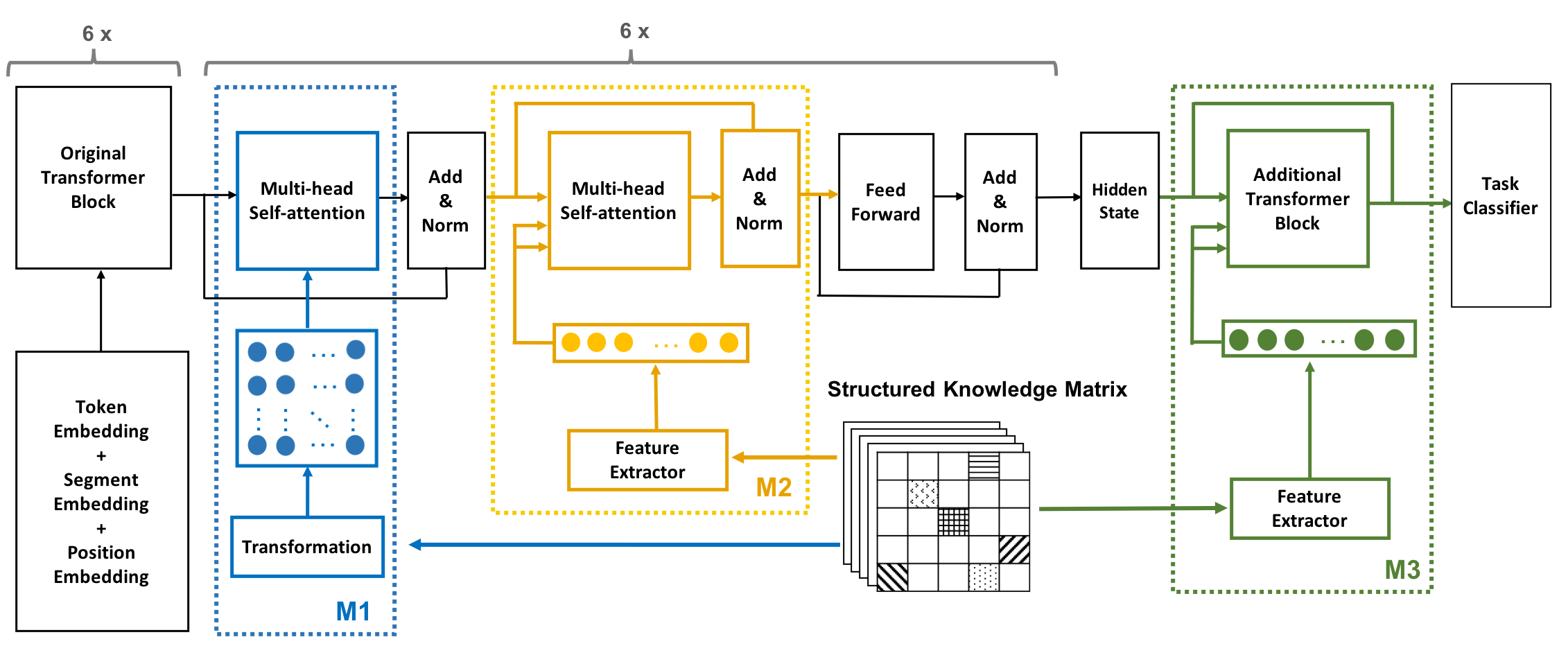}
\caption{An overview of the three different approaches incorporating structured knowledge on $\rm BERT_{BASE}$ (12 $\times$ Transformer). The blue, yellow and green frames denote approaches M1, M2 and M3 respectively.}
\label{fig1}
\end{figure*}

\section{The Models}
\label{sec:models}
We propose three approaches to leverage structured external knowledge in Transformer-based pre-training frameworks.
Figure~\ref{fig1} shows an overview of our models. 
\begin{itemize}
    \setlength{\itemsep}{-1pt}%
    \setlength{\parsep}{-1pt}
    \setlength{\parskip}{-1pt}
    \item \textbf{Structured Knowledge for Attention Weights Adjustment} As shown in the blue block (M1) of Fig~\ref{fig1}, this approach incorporates structured knowledge in  multi-head self-attention layer of Transformer.\\
    \item \textbf{Structured Knowledge for a Separate Knowledge-Specific Layer} As shown in the yellow block (M2) of Fig~\ref{fig1}, this approach adds an additional knowledge-specific multi-head self-attention layer in  Transformer. \\
    \item \textbf{Structured Knowledge for Global Inference}
    As shown in in the green block (M3), this approach directly incorporates structured knowledge into global inference. 
\end{itemize}

We implement all our models based on $\rm BERT_{BASE}$\footnote[1]{\url{https://github.com/google-research/bert}}, and our approaches could be easily extended to other Transformer-based architectures such as GPT.

\begin{figure}[!htpb]
\centering
\includegraphics[width=7.7cm]{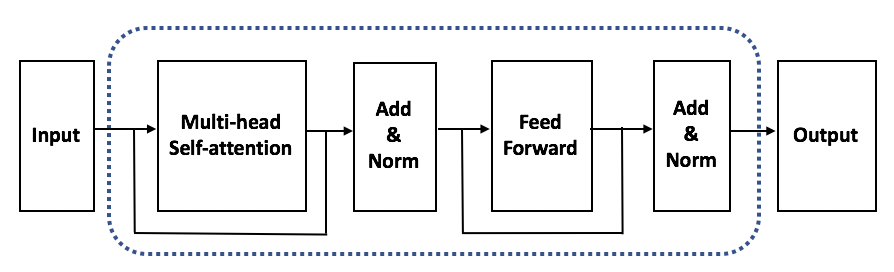}
\caption{Transformer architecture.}
\label{transformer}
\end{figure}

To make the content complete and help introduce our models, we briefly describe Transformer here, and the details can be found in~\cite{vaswani2017attention}. Transformer is widely used as a building block of many state-of-the-art NLP models. Figure~\ref{transformer} shows the architecture of Transformer, which contains a multi-head self-attention layer and a feed-forward layer. Residual connections and normalization are applied between layers. 
$\rm {BERT_{BASE}}$ is built with twelve identical Transformer blocks. The network is randomly initialized and then pre-trained on BooksCorpus~\cite{zhu2015aligning} and a Wikipedia Corpus with two objectives, {\em i.e.}, masked LM and next sentence prediction~\cite{devlin2018bert}.

\subsection{Structured Knowledge}
In general, structured knowledge in relational databases or knowledge graphs can be regarded as a list of triples $\langle n_1, n_2, r \rangle$, where $n_1$ and $n_2$ are nodes and $r$ is their relation. 
In this work, we use triples from commonsense knowledge bases,  WordNet~\cite{miller1995wordnet} and ConceptNet 5.5~\cite{speer2017conceptnet}, as the sources of structured knowledge. 
Although there exist other sources of structured knowledge, {\em e.g.}, Freebase (WikiData)~\cite{bollacker2007freebase}, such knowledge sources mainly convey factual evidence ({\em e.g.}, the relationship between Bill Gates and Microsoft), which is less relevant for general natural language inference. 
A common example of general NLI is a sentence pair like ``a girl in a yellow dress with the sun shining on her face'' and ``a girl in a pink dress with the sun shining on her face'', and determining the relationship between these two sentences relies more on the knowledge of the color ``yellow'' and ``pink'' are two different colors. 

Structured knowledge could benefit natural language inference in two aspects. 
First, external knowledge may in general help align inference-related concepts between a premise and the corresponding hypothesis. For example, the features like \textit{synonymy}, \textit{antonymy}, \textit{hypernymy}, \textit{hyponymy} and \textit{co-hyponyms} can help determine local inference~\cite{chen2017neural}, by helping aligning the mentions of such words between a premise and a hypothesis through attention. 
Second, such semantic relations could be directly used as inference-related features, thus providing auxiliary information for inference. 
For example, knowledge of \textit{hypernymy} and \textit{hyponymy} may help capture entailment, and knowledge of \textit{antonymy} and \textit{co-hyponyms} may help model contradiction.
Considering a sentence pair like ``a man is drinking beer'' and ``a man is drinking whisky'', with the knowledge that ``beer'' and ``whisky'' are different kinds of alcohol (\textit{co-hyponyms}), we can infer that these sentences contradict each other.

In conclusion, given a pair of premise and hypothesis for inference, each word pair $\langle w_i, w_j \rangle$ between the two sentences will be assigned with a multi-dimensional semantic relation $\bm{e_{ij}}$,  $\bm{e_{ij}} \in \mathbb{R}^{k}$, where $k$ represents the dimension of extracted knowledge. Particularly, in the BERT architecture, the premise and hypothesis are concatenated into a sequence of length $n$ (zero-padding shorter sequence or truncate longer sequence to $n$). Therefore, we can form a knowledge matrix $E=\{e_{ij}\}$ with the shape of $n \times n \times k$ for the pair of sentences. 

\subsection{Structured Knowledge for Attention Weights Adjustment}
\label{subsec:method1}
As the structured knowledge can be viewed as a prior to align tokens between premise and hypothesis, we incorporate external knowledge in the multi-head self-attention module of Transformer in the $\rm {BERT_{BASE}}$ architecture. 
The multi-head self-attention module in each Transformer block contains multiple parallel heads and attention weight matrix $A \in \mathbb{R}^{n\times n}$ for each head is calculated by the pair-wise dot product of the transformed input representations, where $n$ represents the length of the input sequence. 

Specifically, we represent the input sequence as $X=[\bm{x_1},\bm{x_2},\dots,\bm{x_t},\dots,\bm{x_n}]$, where $\bm{x_t} \in \mathbb{R}^{d}$ is a continuous representation for a token at position $t$, and $x_t$ is an element-wise summation of token embedding, segment embedding, and position embedding.
In each head, we first calculate 
query ($Q$), 
key ($K$) and
Value ($V$) as below,
which are different transformations of input $X$.
\begin{align}
Q &= [\bm{x_1}, \bm{x_2} ,... ,\bm{x_t}, ... ,\bm{x_n}] W_q \\
K &= [\bm{x_1}, \bm{x_2}, ... ,\bm{x_t} ,... ,\bm{x_n}] W_k \\
V &= [\bm{x_1}, \bm{x_2} ,..., \bm{x_t} ,..., \bm{x_n}] W_v 
\end{align}

The output $H$ as well as the attention weight matrix $A$ are computed as follows:
\begin{align} 
A &=\{a_{ij}\} = softmax(\frac{Q{K}^{\rm T}}{\sqrt{d_k}}) \label{eq1}\\
H &=[\bm{h_1}, \bm{h_2} ,..., \bm{h_t} ,..., \bm{h_n}]^{\rm T} = AV  \label{eq2}
\end{align}
where as in Transformer~\cite{vaswani2017attention}, $\sqrt{d_{k}}$ acts as a scaling factor when calculating the attention weight matrix $A$.

We combine structured knowledge 
$E \in \mathbb{R}^{{n\times n\times k}}$ with self-attention weight $A \in \mathbb{R}^{{n\times n}}$ for each head in the top six Transformer blocks while keeping the bottom six Transformer blocks unchanged. As shown in M1 block of Figure~\ref{fig1}, the structured knowledge matrix is firstly fed into a transformation block to transform into a new matrix $E^\prime = \{e_{ij}^\prime\}$ with the shape $n \times n$ (We choose the average pooling along the last dimension of $E$ for the transformation block).  And then we merge $E^\prime$ into $A$ to get a new attention weight matrix $A^\prime$ as follows:
\begin{equation}
A^\prime = A + A \odot E^\prime  = \{a_{ij} + a_{ij}e_{ij}^\prime\}
\end{equation}
Here $A^\prime$ has the same shape as $A$, and can be directly used to replace $A$ in Equation~\ref{eq2} to get the structured knowledge augmented output $H$. In this way, we inject structured knowledge in each multi-head self-attention layers so as to improve the alignments between tokens.

\subsection{Structured Knowledge for a Separate Knowledge-Specific Layer}
\label{subsec:method2}
Intuitively, the structured knowledge retrieved from human-curated knowledge bases could contain helpful information to determine the relationship between premise and hypothesis.
Therefore, we explore to encode the structured knowledge as knowledge features and then combine these features into the top six Transformer blocks by an additional multi-head self-attention layer shown in M2 of Figure~\ref{fig1}. Also, residual connection and layer normalization are included in this additional layer. 

The extractor used here for knowledge feature extracting is a Convolutional Neural Network (CNN) composed of a stacked convolutional layer with $p$ different filters (Equation~\ref{eq3}) as well as consecutive pooling layers which take its input as the concatenation of the output from convolutional layers (Equation~\ref{eq4}).
The output $H$ of the previous self-attention layer in Transformer block is regarded as query ($Q$) of this additional multi-head self-attention layer, while the extracted knowledge features $C$ are used as its corresponding key ($K$) and value ($V$). Specifically, the output of this newly introduced self-attention module is calculated as below:
\begin{align}
c_i &= \Conv (E, filter_{i}), i \in \{1,\cdots, p\} \label{eq3}\\
C  &= \pooling [c_1;c_2;\cdots;c_p]\label{eq4}\\
A^\prime &= \{a_{ij}^\prime\} = \softmax (\frac{HC^{\rm T}}{\sqrt{d_k}})\label{eq5}\\
P  &= [p_1,p_2,\cdots,p_n]^{\rm T} = A^\prime C\label{eq6}
\end{align}
Where $A^\prime$ and $P$ denote the attention weight and output in the additional multi-head self-attention layer respectively. 
In this way, we incorporate inference-related knowledge to benefit the reasoning.

\subsection{Structured Knowledge for Global Inference}
\label{subsec:method3}
The structured knowledge matrix is expected to provide explicit semantic information to indicate the overall label for a pair of premise and hypothesis. We explore incorporating structured knowledge directly in the global inference layer by adding an additional Transformer block on top of the original $\rm BERT_{base}$ model. This additional Transformer block is a simplified version of a standard Transformer block in BERT with one head instead of multiple ones, and its parameters are randomly initialized and optimized during training. 

The structured knowledge matrix $E$ is first fed into a feature extractor, which generates a structured knowledge matrix $M$ according to Equation~\ref{eq7} to~\ref{eq8}. Similar to the feature extractor used in Section~\ref{subsec:method2}, the feature extractor we use in this approach is also a Convolution Neural Network.
\begin{align}
\bm{m_i} &= \Conv(E,\bm{filter_i}), i \in \{1,\cdots, p^\prime\} \label{eq7}\\
M &= \pooling[\bm{m_1}; \bm{m_2};...; \bm{m_{p^\prime}}] \label{eq8}\\
&w = \softmax (\frac{Mh_0^{\rm T}}{\sqrt{d_k}})\label{eq9}\\
&h_{final} = M\label{eq10}w
\end{align}

Where $h_0$ is the output hidden vector of the first token in output sequence as the compact sentence representation and $h_{final}$ is then fed into the classification layer. As shown in M3 of Figure~\ref{fig1}, for this approach, we combine the structured knowledge matrix $M$ with the compact sentence representation vector $h_0$ by designing an additional Transformer block to produce the input $h_{final}$ for the task classifier. In this way, we manage to leverage structured knowledge in global inference.

\section{Experiment Setup}
\label{sec:experiments}
\subsection{Data}
We evaluate the proposed models on the widely used NLI datasets, {\em i.e.}, SNLI and MultiNLI (MultiNLI-match and MultiNLI-mismatch). We also test our models with the Glockner testset~\cite{glockner2018breaking}. 
These datasets share the same target of a 3-way prediction: determining the relation in a premise-hypothesis pair to be either \textit{entailment}, \textit{neutral}, or \textit{contradiction}. 

For SNLI and MultiNLI, our models are trained and tested based on official splits of training, development, and test set. For the Glockner test set, our models are trained on SNLI.

\subsection{Representation of Structured External Knowledge}
\label{subsec:kb}
The most prominent structured external knowledge for downstream NLI task is lexicon semantics (phrase-level inference knowledge is much more sparse and hard to acquire). 
We use WordNet and ConceptNet as structured knowledge sources and extract specific semantic relations between word pairs.
WordNet is a human-authorized knowledge base in which lexical semantics such as those about nouns, verbs, adjectives, and adverbs are organized into sets of synonyms with semantic relations linking them. ConceptNet5.5 includes massive world and lexical knowledge in different languages from different sources and organizes such knowledge in a graph with differently weighted edges.

In WordNet, we use five types of semantic relation, including \textit{hypernymy, hyponymy, co-hyponyms, antonymy}, and \textit{synonymy}. We extract and represent this knowledge with the method proposed in~\cite{chen2017neural}. 
The dimension $k$ of semantic relation $e_{ij}$ is 5 in our case, if a specific word pair falls into either synonymy or antonymy relation, then the corresponding dimension will be 1, otherwise it will be 0. If two words do not belong to the same synset but share the same hypernym, the value of co-hyponyms dimension will be set as 1, otherwise it is 0. When calculating hypernymy features, it takes the value $(1-n/8)$ if one word is a hypernym of the other word within 8 steps in the WordNet hierarchy. Given calculated hypernymy relation between word $A$ and word $B$, like $[A, B]=0.125$, it is easy to infer that the hyponymy feature is $[B, A]=0.125$. 

We also use ConceptNet 5.5 as a structured knowledge source. The knowledge from ConceptNet will only be added when there is no counterpart in WordNet. 
As mentioned before, the structured knowledge representation will be a 5-dimensional vector. In ConceptNet 5.5, the concepts are aligned into 36 relations. In order to import ConceptNet, we managed to condense ConceptNet relation features compatible with those from WordNet.
Specifically, we filter out relations similar to those selected from WordNet and establish the corresponding relationship between relations from WordNet and ConceptNet as shown in  Table~\ref{concept_condense}. 

Table~\ref{all_pairs} provides the statistics about how many semantic relations for word pairs can be extracted from WordNet and ConceptNet. For ConceptNet, only pairs with both words being English are counted.

\begin{table}[]
\centering
\renewcommand{\arraystretch}{1.2}
\small
\begin{tabular}{p{2cm}|p{4.6cm}}
\toprule
\multicolumn{1}{l|}{\textbf {WordNet}} & \multicolumn{1}{l}{\textbf {ConceptNet}}                                                                             \\ \midrule
Hypernymy                    & HasA                                                                                                       \\ \midrule
Hyponymy                     & \begin{tabular}[c]{@{}l@{}}InstanceOf, Entails, IsA,MannerOf, \\ MadeOf, PartOf, DerivedFrom\end{tabular} \\ \midrule
Co-hyponyms                  & DistinctFrom                                                                                               \\ \midrule
Antonymy                     & Antonym                                                                                                    \\ \midrule
Synonymy        & FormOf, SimilarTo, Synonym                                                                                 \\ \bottomrule
\end{tabular}
\caption{Corresponding relation between WordNet and ConceptNet.}
\label{concept_condense}
\end{table}

\begin{table}[]
\centering
\renewcommand{\arraystretch}{1.2}
\small
\begin{tabular}{p{2cm}|p{2.cm}|p{2.1cm}}
\toprule
            & {\textbf {WordNet}}   & {\textbf {ConceptNet}} \\ \midrule 
Hypernymy   & 753,086   & 5,532      \\
Hyponymy    & 753,086   & 434,381    \\
Co-hyponyms & 3,674,700 & 3,396      \\
Antonymy    & 6,617     & 18,625     \\
Synonymy    & 237,937   & 602,399    \\ \bottomrule
\end{tabular}
\caption{Statistics for semantic relation pairs extracted from WordNet and ConceptNet.}
\label{all_pairs}
\end{table}

\begin{table*}[]
\caption{Results on NLI datasets, comparing our model with previous models. Our model is the combination of the three approaches proposed in this paper. 
The baseline results are from their published papers except those indicated with $\star$
are based on our experiments with the officially released pre-trained models. }
\label{table:overall performance}
\centering
\renewcommand{\arraystretch}{1.0}
\label{table_concept_s}
\begin{tabular}{l|cccc}
\toprule
Model     & SNLI  & MultiNLI-m & MultiNLI-mm & SNLI-Glockner \\ 
\midrule

ESIM~\cite{chen2016enhanced}  & 88.0  & 76.8 & 75.8 & 65.6 \\
KIM~\cite{chen2018neural}     & 88.6  & 77.2 & 76.4 & 83.5\\
ESIM + ELMo
~\cite{Peters:2018} & 89.3 & -  & -  & - \\
GPT~\cite{radford2018improving} & 89.9  & 82.1  & 81.4 & - \\
$\rm {BERT_{BASE}}$~\cite{devlin2018bert}      & 90.6$^\star$  & 84.6   & 83.4    & 94.7$^\star$   \\ 
\midrule

Our Model (w/o ConceptNet) & 90.9 & 84.7  & 84.1   & \textbf{95.3} \\

Our Model (w ConceptNet) & \textbf{91.0} & \textbf{84.9}  & \textbf{84.3}   & \textbf{95.3} \\
\bottomrule
\end{tabular}
\end{table*}

\subsection{Model Details}
\label{subsec:params}
Our model is based on the official pre-trained $\rm {BERT_{BASE}}$.
During fine-tuning, we load all the pre-trained parameters from the officially released pre-trained $\rm {BERT_{BASE}}$ model, and then we initialize other parameters, including those for the task-specific classifiers and those we newly introduced in our model. $\rm {BERT_{BASE}}$ consists of 12 Transformer building blocks, in which the number of heads for the attention module is 12, and the size of the hidden state is 768. The maximum length of the input sequence is 128.

The feature extractor we used in Section~\ref{subsec:method2} is CNN, with a 4-layer stacked convolutional layer and following two pooling layers. The filters for the stacked convolutional layer have kernel sizes of $3\times3$, $5\times5$, $7\times7$, and $9\times9$. The number of channels for each layer is 16 with a stride of 1. After feeding the same structured knowledge matrix into different convolutional layers, we concatenated the four outputs by the last dimension, and then it is followed by 2 max pooling layers, which have pooling sizes of $2\times2$ and $5\times5$, and are applied at stride 2 and 3, respectively. There is a single-layer feed forward network at the end to transform the structured knowledge vector into an appropriate size for further processing.

The feature extractor we used in Section~\ref{subsec:method3} is a CNN with a similar architecture as the CNN used in Section~\ref{subsec:method2}. One of the different points is that it includes a 3-layer stacked convolutional layer instead of a 4-layer stacked one, and the filters are with the shape of $3\times3$, $5\times5$, and $7\times7$. The other difference is mainly about the parameters of the single-layer feed forward network since we need to transform the knowledge vector into different space with a different dimension. 

\section{Experiment Results}
\label{sec:results}

\subsection{Overall Performance}
Table~\ref{table:overall performance} presents the results of our model on SNLI, MultiNLI-matched, MultiNLI-mismatched, as well as the Glockner test set.
Our model here represents the overall model that combines M1, M2, and M3 discussed before.
We first test our model with WordNet while without structured knowledge from ConceptNet. It shows that even using only WordNet can bring improvements compared with the baseline model. The best performance was achieved by leveraging knowledge from both WordNet and ConceptNet.
Specifically, our model with the complete set of structured knowledge outperforms the baselines consistently across all these datasets compared with $\rm BERT$.
It shows that leveraging structured knowledge on pre-trained models can benefit NLI tasks. 
We have not experimented with ${\rm BERT_{LARGE}}$ because it is computationally expensive, but we believe the proposed model is orthogonal and useful, particularly when structured knowledge is complementary to knowledge learned in pretraining, e.g., when domain-specific tasks are concerned with abundant structured domain knowledge available.

\subsection{Ablation Analysis}
\begin{table*}[]
\caption{Accuracies of M1, M2, and M3 on SNLI, MultiNLI-m, MultiNLI-mm, and Glockner. M1, M2 and M3 refer to the submodel in Section~\ref{subsec:method1},~\ref{subsec:method2}, and~\ref{subsec:method3}, respectively. The resource of structured knowledge is WordNet, and the models tested on Glockner are trained with SNLI.}
\label{table:M123}
\centering
\renewcommand{\arraystretch}{1.0}
\label{table_concept_s}
\begin{tabular}{l|cccc}
\toprule
Model     & SNLI  & MultiNLI-m & MultiNLI-mm & SNLI-Glockner \\ 
\midrule
${\rm BERT_{BASE}}$~\cite{devlin2018bert}      & 90.6  & 84.6   & 83.4    & 94.7   \\ 
\midrule

M1 (w/o ConceptNet)     & 90.7 & \textbf{85.0}  & 84.1   & 95.0\\
M2 (w/o ConceptNet)     & 90.7 & 84.6  & \textbf{84.2}   & 95.0  \\
M3 (w/o ConceptNet)     & \textbf{90.9} & \textbf{85.0}  & 83.9   & \textbf{95.6}  \\ 
\bottomrule

\end{tabular}
\end{table*}

\begin{figure}[t]        
 \center{\includegraphics[width=8cm]  {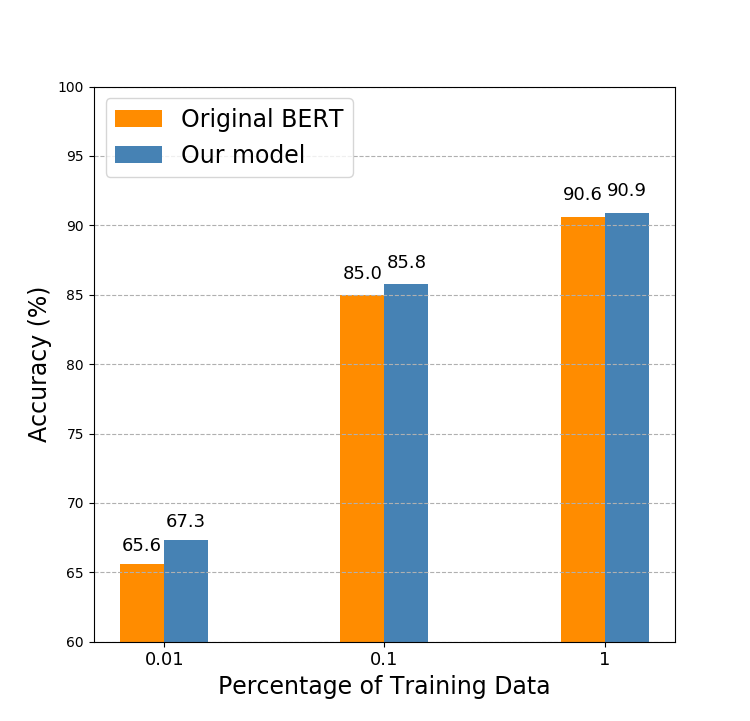}}        
 \caption{\label{figA} 
 Accuracies of original BERT and our structured knowledge (with WordNet) enhanced model with different proportion of training data on SNLI.
 }     
\end{figure}

To further prove that structured external knowledge can benefit BERT on NLI, we perform ablation tests to get empirical results for analysis. 

The first test aims to observe the performance of M1, M2, and M3, respectively. The performances of three different approaches are shown in Table~\ref{table:M123}. Comparing with the results from $\rm BERT_{BASE}$, we can see that every single method can bring improvements on these datasets to different degrees.  

In Figure~\ref{figA}, we present the test accuracy from models trained with/without structured external knowledge. 
A comparison of both models under each data split shows that the model with structured knowledge performs consistently better than the original BERT baseline.
The gap between the two models becomes larger when less training data is available.
When we have only 0.01 percent of the data available, structured knowledge has more obvious effects on model performance. 
Our model shows extra benefit from the structured relation pairs, complementary to information from training set and pre-training.

\begin{figure}[]        
 \center{\includegraphics[width=8cm]  {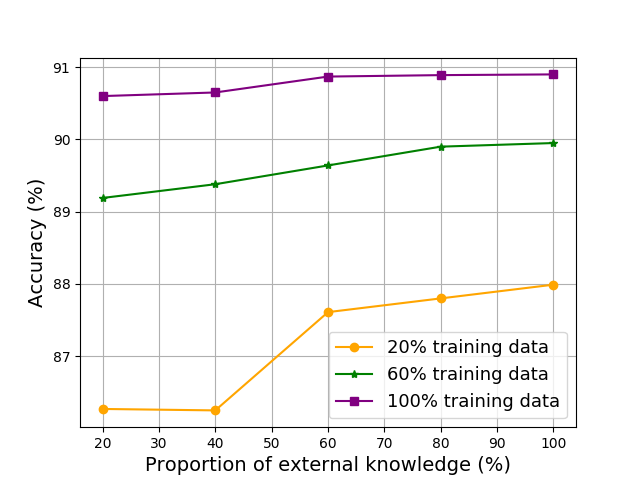}}        
 \caption{\label{figB} 
 Accuracies of the overall model leveraging different proportion of structured knowledge (WordNet), and trained with different proportion of SNLI training data.
 }     
\end{figure}

Figure \ref{figB} shows increasing structured knowledge has a positive effect on test accuracy. The slope of each single line in Figure \ref{figB} shows that when we increase the amount of structured knowledge from 20\% to 100\%, the test accuracy increases accordingly, and the accuracy increases more obviously when there is less training data available.
This increasing trend provides evidence that the proposed model is capable of using lexical-level structured knowledge pairs for natural language inference. 

\begin{table}[ht]
\renewcommand{\arraystretch}{1.0}
\begin{tabular}{p{6.2cm}|p{0.6cm}}
\toprule
Premise-Hypothesis pairs & P~/~T\\ \midrule
\textit{\textbf {p:}} A man playing an electric \textbf{\color{blue} guitar} on stage. \\
\textit{\textbf {h:}} A man playing \textbf{\color{blue} banjo} on the floor. & E~/~C\\ \midrule
\textit{\textbf {p:}} A man jiggles bowling pins for the \textbf{\color{blue} first} time.\\
\textit{\textbf {h:}} A man jiggles bowling pins for the \textbf{\color{blue} 1st} time. & C~/~E\\ \midrule
\textit{\textbf {p:}} A woman is sipping some \textbf{\color{blue}wine}. \\
\textit{\textbf {h:}} A woman is sipping some \textbf{\color{blue}vodka}. & N~/~C\\ 
\bottomrule
\end{tabular}
\caption{Examples from Glockner and SNLI on which our model predicts correctly while not the original BERT. P~/~T represents prediction and true label. E, N, and C refer to entailment, neutral and contradiction, respectively.}
\label{K_BERT}
\end{table}

\subsection{Case Study}
To further find out how structured knowledge enhances the pre-trained BERT, we analyze some specific cases from SNLI and Glockner, on which only our model can yield correct predictions but not original BERT, as shown in Table~\ref{K_BERT}. Note that all the highlighted keyword pairs can be retrieved in the structured knowledge base we use. Take the first pair of sentences in Table~\ref{K_BERT} as an example, the highlighted pair of words ``guitar'' and ``banjo'' in the first example can be retrieved in WordNet with the relation of \textit{co-hyponyms}, which indicates they have the same hypernym ``instrument'' but refer to different entities. Our model can leverage this kind of knowledge to make a reasonable inference that these two sentences contradict to each other as this kind of lexical semantic knowledge is too sparse to learn from unstructured training data. 

\section{Conclusions and Discussion}
\label{sec:conclusion}
Research on natural language inference has considerably benefited from large annotated datasets.
Most recently, there have been two lines of approaches to employing knowledge available outside training data: knowledge from structured knowledge bases and that learned from unsupervised pretraining. In this paper, we explore whether these two approaches complement each other, and how to develop models that can bring together their advantages. We propose models that leverage structured knowledge in different components of pre-trained models. The results show that the proposed models perform better than the BERT baseline. While our models are proposed for NLI, they can be easily extended to other sentence or sentence-pair classification problems to leverage these two sources of external knowledge.


\bibliography{anthology,custom}
\bibliographystyle{acl_natbib}

\end{document}